\documentclass[a4paper,fleqn]{cas-sc}

\usepackage[authoryear,longnamesfirst]{natbib}
\usepackage{algorithm}
\usepackage{algorithmic}
\usepackage{multirow}
\usepackage{enumitem}

\def\tsc#1{\csdef{#1}{\textsc{\lowercase{#1}}\xspace}}
\tsc{WGM}
\tsc{QE}
\tsc{EP}
\tsc{PMS}
\tsc{BEC}
\tsc{DE}
\begin{document}
\let\WriteBookmarks\relax
\def\floatpagepagefraction{1}
\def\textpagefraction{.001}
\shorttitle{Leveraging social media news}
\shortauthors{CV Radhakrishnan et~al.}
%\begin{frontmatter}

\title [mode = title]{MetaConcept: Learn to Abstract via Concept Graph for
	Weakly-Supervised Few-Shot Learning}                      
%\tnotemark[1,2]

%\tnotetext[1]{This document is the results of the research
%   project funded by the National Science Foundation.}

%\tnotetext[2]{The second title footnote which is a longer text matter
%   to fill through the whole text width and overflow into
%   another line in the footnotes area of the first page.}

\author[1]{Baoquan Zhang}[style=chinese]
%\cormark[1]
%\fnmark[1]
%\ead{baoquanzhang@yeah.net}
%\ead[url]{www.cvr.cc, cvr@sayahna.org}

%\credit{Conceptualization of this study, Methodology, Software}

\address[1]{Department of Computer Science and Technology, Harbin Institute of Technology, Shenzhen, China}

\author[1]{Ka-Cheong Leung}[style=chinese]

\author[1]{Yunming Ye}[style=chinese]
\cormark[1]
%\fnmark[2]
\ead{yeyunming@hit.edu.cn}
%\ead[URL]{www.sayahna.org}

%\credit{Data curation, Writing - Original draft preparation}

\author%
[1]
{Xutao Li}[style=chinese]
%\cormark[2]
%\fnmark[1,3]
%\ead{rishi@stmdocs.in}
%\ead[URL]{www.stmdocs.in}

\cortext[cor1]{Corresponding author.}
%\cortext[cor2]{Principal corresponding author}
%\fntext[fn1]{This is the first author footnote. but is common to third
%  author as well.}
%\fntext[fn2]{Another author footnote, this is a very long footnote %and
%  it should be a really long footnote. But this footnote is not yet
%  sufficiently long enough to make two lines of footnote text.}

%\nonumnote{This note has no numbers. In this work we demonstrate &$a_b$
%  the formation Y\_1 of a new type of polariton on the interface
%  between a cuprous oxide slab and a polystyrene micro-sphere placed
%  on the slab.
%  }

\begin{abstract}
Meta-learning has been proved to be an effective framework to
address few-shot learning problems. The key challenge is how to
minimize the generalization error of base learner across tasks. In
this paper, we explore the concept hierarchy knowledge by leveraging
concept graph, and take the concept graph as explicit meta-knowledge for the base learner,
instead of learning implicit meta-knowledge, so as to boost the classification performance of meta-learning on weakly-supervised few-shot learning problems. To this end, we propose a
novel meta-learning framework, called MetaConcept, which learns
to abstract concepts via the concept graph. Specifically, we firstly
propose a novel regularization with multi-level conceptual abstraction
to constrain a meta-learner to learn to abstract concepts via
the concept graph (i.e. identifying the concepts from low to high
levels). Then, we propose a meta concept inference network as
the meta-learner for the base learner, aiming to quickly adapt to a
novel task by the joint inference of the abstract concepts and a few
annotated samples. We have conducted extensive experiments on
two weakly-supervised few-shot learning benchmarks, namely, WSImageNet-
Pure and WS-ImageNet-Mix. Our experimental results
show that 1) the proposed MetaConcept outperforms state-of-the-art
methods with an improvement of 2\% to 6\% in classification
accuracy; 2) the proposed MetaConcept can be able to yield a good
performance though merely training with weakly-labeled data sets.

\end{abstract}


\maketitle

\section{Introduction}
Few-Shot Learning (FSL) is a machine learning approach for understanding new concepts with a few examples. It targets at acquiring good learning performance by leveraging the prior knowledge for a novel task where its class is unfamiliar and only a little supervised information is available \cite{jamal2019task, li2019finding, wang2019few}. The study of FSL has received much attention recently because of the following features: 1) FSL is a cheap learning paradigm, which can reduce the costs of data annotations for many data-dependent applications, such as image classification \cite{jamal2019task, li2019finding, rusu2018meta}, object detection \cite{dong2018few, fu2019meta, kang2019few}, and neural architecture search \cite{brock2017smash, liu2019metapruning}. 2) FSL can be directly applied to rare case learning applications, where the acquisition of annotated samples is hard or impossible due to scarcity or safety concerns, such as cold-start item recommendation \cite{vartak2017meta} and drug discovery \cite{altae2017low}.  

\begin{figure}
	\begin{center}
		\includegraphics[width=1.0\linewidth]{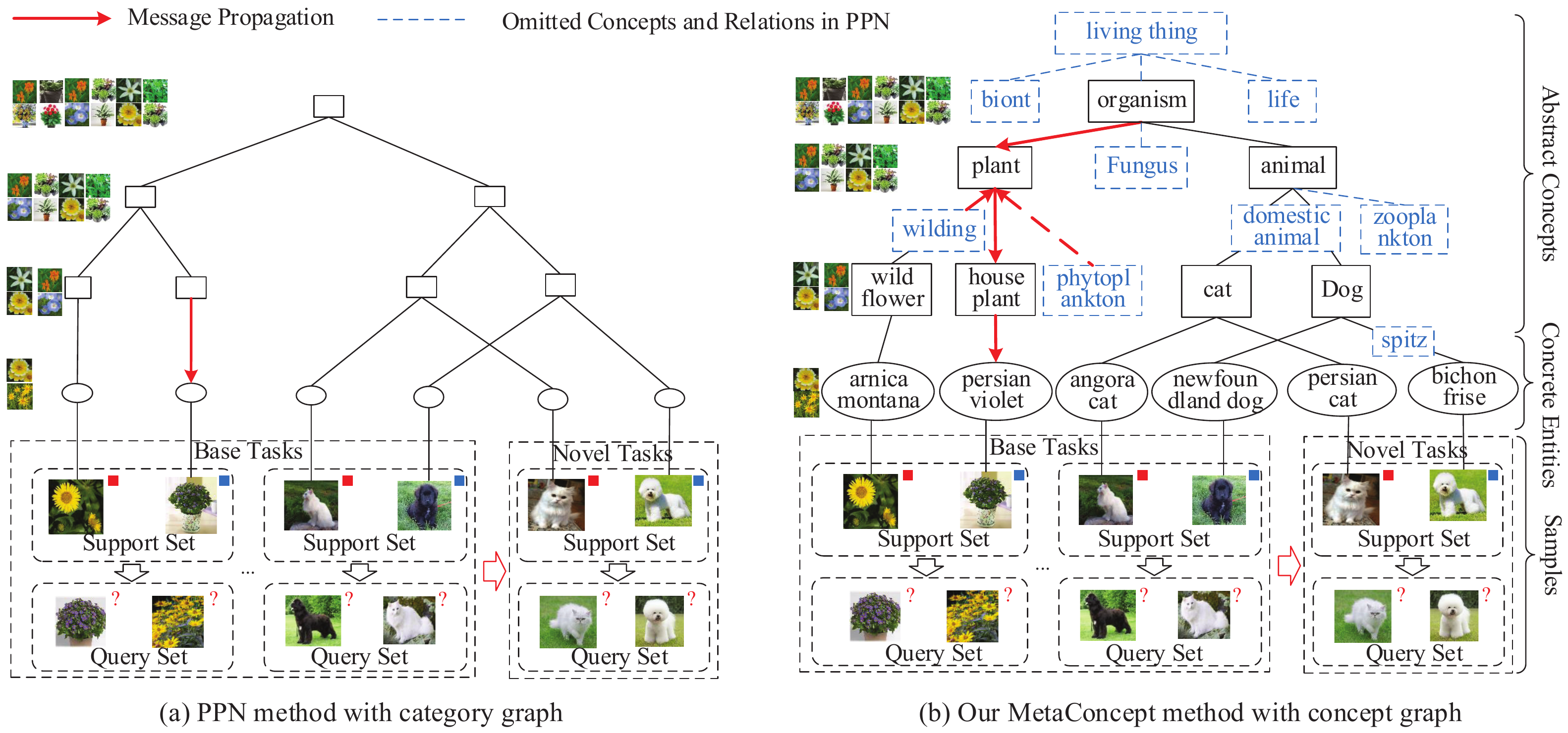}
	\end{center}
	\caption{Illustration of our motivation by the comparison of PPN and our MetaConcept methods. We depict an example of few-shot classification task with two categories  to shows the key ideas. There is only a labeled sample for each category. Here, the concept graph is obtained from WordNet and we only show the message propagation paths of the node `Persian violet' on category graph and concept graph, respectively.}
	\label{fig:motivation}
\end{figure}

At present, most methods primarily focus on meta-learning frameworks to solve the FSL problems \cite{jamal2019task, snell2017prototypical, finn2017model, finn2018probabilistic, nichol2018reptile, andrychowicz2016learning, ravi2016optimization}. They aim to learn a base learner based on meta-knowledge from past experience so as to quickly adapt to novel tasks by just a few annotated samples. Specifically, the framework consists of two major phases: 1) learning meta-knowledge from base tasks sampled from the same distribution (called meta-training phase); and 2) quickly constructing or fine-tuning a base learner by employing the learned meta-knowledge and a few annotated samples to the task-specific model for novel tasks (called meta-test phase). Therefore, what meta-knowledge to learn determines the level of the generalization performance of the base learner across tasks. Generally speaking, the meta-knowledge is explored by a neural network to be treated as a black box without any prior information. It can be a common initialization \cite{finn2017model, finn2018probabilistic, nichol2018reptile}, a shared metric \cite{sung2018learning}, an universal optimization \cite{andrychowicz2016learning, ravi2016optimization}, or a generic embedding network \cite{snell2017prototypical, liu2019prototype}. For example, in \cite{ravi2016optimization}, an LSTM without any prior information is adopted as meta-knowledge, aim to learn a optimization algorithm on finely-labeled data sets, so that the optimization algorithm can quickly train a base learner when only a few labeled samples is available. Currently, these methods have shown the superior performance in solving the FSL problem. However, most methods just focus on learning implicit meta-knowledge on finely-labeled data, ignoring explicit prior knowledge (e.g. concept graph) and weakly-labeled data which is cheap to collect. 

Recently, Liu et al. \cite{liu2019prototype} has explored weakly-supervised information for FSL and define a novel FSL problem called Weakly-Supervised FSL (WSFSL). Specifically, they proposed a Prototype Propagation Networks (PPN), aiming to obtain a more robust class prototypes by aggregating the messages (prototypes) from all the parent classes on category subgraph. Their experimental results shown that it is helpful for boosting classification performance of meta-learning on few-shot classification. However, the WSFSL problem has not been further explored in recent years. As shown in Figure 1, we find that the PPN method still has the following limitations: 1) they just explore the graph structure informtation, ignoring the semantic information of graph node (i.e.\ category semantics) which is useful for distinguishing categories with similiar graph structure; 2) the category graph is extracted from knowledge graph (e.g. WordNet) according to all categories of datasets, which usually filter out a large intermediate or irrelevant abstract concepts so as to obtain clear category hierarchy \cite{liu2019prototype}. However, the ignored abstract concepts and relations (marked in the blue dotted line) still contains abundant prior information used to exploring category hierarchy; and 3) the prototype propagation strategy performed on category subgraph just explore first-order information (shown in the red line in Figure 1(a)) of graph, ignoring high-order information (shown in the red line in Figure 1(b)) which is helpful for learning robust node (or category) representation. Thus, the concept hierarchy is not fully explored for the WSFSL problem, which limits the classification performance of meta-learning on the WSFSL problem.

In this paper, we focus on WSFSL problem \cite{liu2019prototype} and propose a novel concept graph-based meta-learning framework (known as MetaConcept) towards the limitations mentioned above. Specifically, we introduce a concept graph \cite{ji2019microsoft, miller1995wordnet} as explicit meta-knowledge and propose a cross-level meta-learner to fully explore the concept hierarchy knowledge for WSFSL. Different from the category graph used in PPN, the concept graph contains more abundant abstract concepts, relations, and semantic information. Our experimental results show that it can further boost the classification performance of meta-learning on WSFSL, around 3\% to 12\% in classification accuracy. Specifically, the idea is inspired by the basic-level categorization \cite{wang2015inference}, i.e.\ people can understand an unfamiliar object by mapping it into an appropriate level of concepts \cite{ji2019microsoft}. In Figure~\ref{fig:motivation}(b), we depict an example to illustrate the key idea. The concept graph consists of abstract concept and concrete entity levels, which offers an explicit transfer manner for the base learner to adapt from the base tasks to novel tasks, because of the following features: 1) the disjoint classes can share common abstract concepts. For example, the classes of ``Newfoundland dog'' and ``Bichon frise'' are disjoint at the concrete entity level, but they share the same set of concepts ``dog'', ``dimestic animal'', ``animal'', ``organism'', and ``living thing'' at the abstract concept level; 2) the entities/concepts can be understood and represented in a common semantic space by a large unsupervised text corpora; and 3) the weakly-labeled data sets are collected from multiple levels, which is helpful for exploring the concept hierarchy and learning a robust cross-level knowledge inference strategy. 

Based on this idea, in MetaConcept, we propose a novel meta-learning framework consisting of a multi-level conceptual abstraction-based regularization and a meta concept inference network. Here, the former is a regularizer for constraining the latter meta-learner. Specifically, we explore the concept hierarchy in two ways: 1) for each task, we explore the multi-hops relation of selected classes on concept graph (as shown in Figure~\ref{fig:motivation}(b)) by introducing graph convolutional networks;  2) we construct multi-level auxiliary tasks according to the weakly-labeled data sets, to train the meta-learner to infer FSL classifiers at any level. During training, we firstly sample batches of few-shot classification tasks from each level of datsets, divided into few-shot entity and concept classification task according to their level on concept graph, which is termed as a episode \cite{vinyals2016matching}. Different from the PPN method, these tasks sampled from different levels are independent of each other, aiming to explore concept hierarchy on global graph, instead of subgraph. And then we taking the concept graph as inputs of the meta-learner to infer task-specific classifiers which depends on the selected classes (nodes of concept graph) for each few-shot classification task. Finally, we make use of these tasks to train the cross-level meta-learner, so that the meta-learner is able to infer not only an entity classifier but also a concept classifier at different levels (i.e.\ learn to abstract). Here, we take the vaild loss on few-shot concept classification tasks as a regularization, aiming to boost the classification performance on few-shot entity classification tasks. During test, different PPN+ method proposed in \cite{liu2019prototype}, the MetaConcept method directly performs few-shot classification on novel classes which does not require weakly-supervise information annotations. Our experimental results show that the proposed MetaConcept method exploring explicit concept hierarchy knowledge can significantly boost the classification performance of the base learner on few-shot entity classification tasks. The main contributions of this work have three-fold: 
\begin{itemize}[]
	\setlength{\itemsep}{0pt}
	\setlength{\parsep}{0pt}
	\setlength{\parskip}{0pt}
	\item We explore concept hierarchy knowledge by leveraging concept graph for WSFSL. Here, we take the concept graph as explicit meta-knowledge, instead of learning implicit meta-knowledge, so as to boost the classification performance of meta-learning on WSFSL.
	\item We propose a concept graph-based meta-learning framework consisting of a multi-level conceptual abstraction-based regularization and a meta concept inference network which explores the high-order infomation of concept graph. Here, the former is a regularizer for constraining the latter meta-learner, aiming to learn a universal meta-learner for enabling the base learner quickly adapts to novel tasks at any level.
	\item We have conducted extensive experiments on two realistic datasets, namely, WS-ImageNet-Pure and WS-ImageNet-Mix. Our experimental results show that 1) the proposed MetaConcept can improve the classification performance on novel tasks by 2\%-6\% in accuracy, and 2) the proposed MetaConcept is able to achieve good performance when it is only trained on weakly-labeled data sets.
\end{itemize}

The rest of this paper is organized as follows. In Section 2, we have a brief survey on the related work of the FSL. In Section 3, we propose our concept graph-based MetaConcept framework. In Section 4, we validate our methods and make comparisons with other approaches on three realistic datasets. In Section 6, we come to a conclusion and discuss the future work.

\section{Related Work}
In this section, we briefly summarize related work into two categories: (1) Graph Neural Networks, (2) Zero-Shot Learning, and (3) Meta-Learning.

\subsection{Graph Neural Network}
Graph neural network (GNN) is a type of the deep neural network, which offers a connectionist model for learning from graph-structured data end-to-end \cite{wu2019comprehensive}. Recently, the GNN has drawn a vast interest in various domains, including social network \cite{fan2019graph, zhang2018link}, knowledge graph \cite{lin2019kagnet, wang2018cross}, and computer vision \cite{johnson2018image, qi2018learning}. Graph convolution network (GCN) is one of the classical methods in this family. In \cite{kipf2016semi}, the GCN has been employed for solving semi-supervised graph learning problems. It adopts a local graph convolution to represent the current node by aggregating its neighboring nodes, aiming to acquire more robust graph representation. Here, GCN has two key advantages. First, it can learn a good low-dimension embedding for node and graph from the network structure and node information. Second, it can explicitly extract multi-hop representation through node message aggregation layer-by-layer. Hence, we adopt the GCN framework to model a meta-learner aiming to learn a robust abstract and inference strategy on the concept graph for solving the WSFSL.

\subsection{Zero-Shot Learning}
Zero-shot learning (ZSL) is closely related to FSL, whose objective is to recognize an unseen category when no supervision information is available \cite{yongqin2019zeros, chenrui2019transferable}. The key idea is to build semantic connections between the seen classes and unseen classes by exploiting and exploring the prior knowledge. Previous studies mainly focus on semantic embedding-based approaches to address the ZSL problem, which learn a transferable projection function between visual features and semantic representations from the auxiliary data \cite{elyor2017semantic}. Recently, a graph based approaches are developed for ZSL. In \cite{michael2019rethinking} and \cite{xiao2018zeros}, a knowledge graph is introduced to build classifier predictor. In \cite{zhimao2019few}, the idea is further extended to FSL and a two-stage training framework is built. Though our method is also graph based, there are two key differences from the previous studies: 1) we incorporate weakly-supervised information at multi-levels to fully exploit the concept hierarchy knowledge; 2) we propose a novel meta-learning framework with concept graph, which works in an end-to-end manner.

\subsection{Meta-Learning}
Meta-learning has been proved to be an effective method for solving the FSL \cite{lee2019meta, mishra2017simple, qiao2018few}. Many meta-learning methods have been proposed from various approaches such as metric-based approaches \cite{chen2019closer, snell2017prototypical, sung2018learning, vinyals2016matching}, optimization-based approaches \cite{finn2017model, finn2018probabilistic, jamal2019task, li2019finding, nichol2018reptile}, and graph-based approaches \cite{kim2019edge, liu2019prototype, liu2019learning, satorras2018few}. 

Metric-based approaches follow a simple nearest neighbour framework and aim at learning a common metric space shared with different tasks by minimizing intra-class similarity while maximizing the similarity between different classes. ProtoNet \cite{snell2017prototypical} makes use of the euclidean-based distance as a similarity measure among samples, where they make use of the similarity of query samples with the prototype of support samples belonging to same class to predict the probability of each class. Based on the simplicity of ProtoNet, the AM3 network \cite{xing2019adaptive} introduces novel semantic information to boost the robustness of the prototype for each class.

Optimization-based approaches aim to learn a effective initialization and optimization method across different tasks. MAML \cite{finn2017model} is a typical work in this family, which aims to learn an effective initial parameter for a base learner, so that the base learner can generalize well to novel tasks by a few fine-tuning steps. Based on the idea, many methods extend this work such as Reptile \cite{nichol2018reptile}, LEO \cite{rusu2018meta}, and Probabilistic MAML \cite{finn2018probabilistic}. For example, Reptile has proposed an extended MAML that do not need to unroll a computation graph, making MAML faster in computation \cite{nichol2018reptile}. 

Graph-based approaches follow from the GNN frameworks, aiming to solve the FSL problems by the supervised message passing networks. For example, a GNN being trained end-to-end has been proposed in \cite{satorras2018few}, where the nodes are associated with images, and edges are given by a trainable similarity kernel for few-shot classification tasks. In \cite{liu2019learning}, a novel transductive propagation network was devised for FSL, targeting at learning to propagate labels from support samples to query samples. In \cite{liu2019prototype}, it introduces a novel graph structure defined on prototype levels, and proposes a prototype propagation network for WSFSL. This aims at propagating the prototype of classes on a subgraph sampled from the graph structure for few-shot classification tasks. 

Our proposed technique can be considered as a combination of the graph-based and optimization-based approaches. Yet, it differs with existing methods in three ways. First, we adopt the concept graph as explicit meta-knowledge of the base learner, instead of learning implicit meta-knowledge, so as to minimize the generalization error across tasks. Second, we model a cross-modal and universal meta-learner via the concept graph, aiming to inferring FSL classifier at any level. Finally, our method focuses on using a global concept graph, not on a subgraph. This can enhance the performance of the base learner on novel tasks by fully exploiting concept hierarchy knowledge on the global concept graph.

\section{Methodology}

For the FSL problem, it is difficult to learn a robust deep model by exploring only a little of supervised information. Fortunately, the weakly-labeled data and explicit prior knowledge are usually free or cheap to collect. In this paper, we focus on WSFSL and propose a novel meta-learning framework to explore explicit concept hierarchy knowledge by leveraging the two types of information. 

%-------------------------------------------------------------------------
\subsection{Preliminaries and Notation}
\label{section:3_1}

Formally, given three finely-labeled data sets: a training set $D^{tr}$ with a set of classes $C^{tr}$ (i.e.\ meta-training class set), and two data sets (a support set $D^{su}$ and a test set $D^{te}$) sharing the same label space with a set of classes $C^{te}$ (i.e.\ meta-test class set). Here, the sets $C^{te}$ and $C^{tr}$ are disjoint, called target entity set. Furthermore, we construct a concept graph according to the target entity set via the hierarchical relation of categories/concepts. In the concept graph, a leaf node denotes a concrete entity. A non-leaf node corresponds to an abstract concept or a coarse class. An edge represents an abstract relationship between two abstract concepts, as shown in Figure~\ref{fig:motivation}(b). Formally, the concept graph $G=(V,E)$ include $N^{le}$ abstract levels, $M$ nodes $v_i \in V$, a number of edges $(v_i, v_j) \in E$, a binary adjacency matrix $A \in \mathbb{R}^{M \times M}$, a degree matrix $D_{ii}= \sum_{j} A_{ij}$, and a $d$-dimension concept semantic embedding $Z\in \mathbb{R}^{M \times d}$. In addition, we assume that there exists a set of weakly-labeled data $D^{we}=\{D^{we}_{l}\}_{l=0}^{N^{le}-1}$ with a set of coarse classes $\{C^{we}_{l}\}_{l=0}^{N^{le}-1}$ because it is usually free or cheap to collect, where $l$ denotes the abstract level of the concept graph. Our aim is exploring the concept graph and weakly-labeled data to address the FSL problem. That is, we need to learn a classifier via the train set $D^{tr}$, weakly-labeled data set $D^{we}$, and concept graph $G$ for the test set $D^{te}$ with unseen classes $C^{te}$, for which only a few labeled examples are available in the support set $D^{su}$. The problem is called $N$-way $K$-shot problem when the test set $D^{te}$ includes $N$ unseen classes and each class in $D^{su}$ contains $K$ labeled samples.

Specifically, in meta-training phase, we mimic the setup of the sets $D^{su}$ and $D^{te}$, and construct a large number of tasks from the training set $D^{tr}$, called ``few-shot entity classification tasks''. Here, each task $\tau$ consists of $N$ classes (i.e.\ $\{C_{i}\}_{i=0}^{N-1}$) sampled from $C^{tr}$, and includes a support set $S=\left\{(x_i, y_i)\right\}_{i=0}^{m-1}$ ($m=N \times K$) with $K$ labeled samples from each of the $N$ classes and a query set $Q=\left\{(x_i, y_i)\right\}_{i=0}^{n-1}$. We then perform meta-learning on the few-shot entity classification tasks to explore transferrable knowledge. Therefore, the estimation of likelihood maximization for our meta-learning based on the concept graph can be written as:

\begin{equation}
\max\limits_{\theta} \ \mathbb{E}_{\tau \sim T^{e}} \ \left[\mathbb{E}_{S,Q \sim \tau} \sum_{(x,y) \in Q} log(P(y|x, S, G, D^{we}, \theta ))\right]
\label{con:3_1_1}
\end{equation}
where $T^{e}$ denotes a set of few-shot entity classification tasks. For clarity, the notations mentioned above are summaried in Table~\ref{table:notations}.

\begin{table}[width=.9\linewidth,cols=4,pos=h]
	\caption{A summary of the notations used in the paper.}\label{table:notations}
	\begin{tabular*}{\tblwidth}{@{} LLLL@{} }
		\toprule
		Notation & Definition & Notation & Definition\\
		\midrule
		$D^{tr}$ & training set & $G$ & concept graph \\
		$D^{su}$ & support set & $V$ & node set \\
		$D^{te}$ & test set & $E$ & edge set \\
		$D^{we}$ & weakly-labeled data set & $N^{le}$ & number of abstract levels \\
		$C^{tr}$ & meta-training classes set & $M$ & number of node \\
		$C^{te}$ & meta-test classes set & $A$ & binary adjacency matrix \\
		$C^{we}$ & coarse classes set & $D$ & degree matrix \\
		$S$ & support set of base tasks & $Z$ & concept semantic embedding \\
		$Q$ & query set of base tasks &  &  \\
		$N$ & number of classes of few-shot classification task & $-$ & - \\
		$K$ & number of labeled samples of each class &  &  \\
		\bottomrule
	\end{tabular*}
\end{table}
%-------------------------------------------------------------------------

\subsection{Multi-Level Conceptual Abstraction}
\label{section:3_2}
The key challange of the problem defined in Eq. (\ref{con:3_1_1}) is how to learn to abstract concepts via the concept graph $G$ and the weakly-labeled data set $D^{we}$ (i.e.\ exploring the concept hierarchy knowledge). The challenge can be addressed by the multi-level concept classification. That is, we take each abstract concept node as a class and apply the concept classification at each abstract level $l=0,1,...,N^{le}-1$. We name the process as multi-level conceptual abstraction (MLCA).

Specifically, we construct multi-level auxiliary tasks from abstract concept levels, called ``few-shot concept classification tasks''. The setting of the few-shot concept classification task is similiar to the few-shot entity classification tasks defined in Section \ref{section:3_1}. The difference is that 1) the class set of each few-shot concept classification task is sampled from the set of abstract concepts (coarse classes), i.e.\ non-leaf nodes of the concept graph; and 2) the samples $(x,y)$ of each task at level $l$ are taken from the weakly-labeled data $D^{we}_l$. Then, the few-shot concept classification task is performed in forms of regularization in the meta-training phase. The regularization aims to constraint a meta-learner to infer not only an entity classifier but also a multi-level concept classifier. To this end, the estimation of likelihood maximization of our meta-learning based on the concept graph can be further expressed as:

\begin{equation}
\begin{aligned}
\max \limits_{\theta} \ \lambda_e \mathbb{E}_{\tau \sim T^{e}} \left[\mathbb{E}_{S,Q \sim \tau} \sum_{(x,y) \in Q} log(P(y|x, S, G, \theta ))\right] + \lambda_c \sum_{l=0}^{N^{le}-1}\mathbb{E}_{\tau \sim T^{c}_{l}} \left[\mathbb{E}_{S,Q \sim \tau} \sum_{(x,y) \in Q} log(P(y|x, S, G, \theta ))\right]
\end{aligned}
\label{con:3_2_1}
\end{equation}
where $T^{c}_{l}$ denotes the set of few-shot concept classification tasks at the abstract level $l$, and $\lambda_e$ and $\lambda_c$ are hyperparameters adjusting the weight of regularization. Following the setting of WSFSL in [22], $\lambda_e$ is set to one by default. In particular, the learning problem becomes more economical when $\lambda_e$ is set to zero, because the meta-learner is trained merely on the weakly-labeled data set $D^{we}$, which is much cheaper to obtain than finely-labeled data.

%-------------------------------------------------------------------------
\subsection{Meta Concept Inference Network}
\label{MCIN}

\begin{figure*}
	\begin{center}
		%\fbox{\rule{0pt}{2in} \rule{.9\linewidth}{0pt}}
		\includegraphics[width=0.9\linewidth]{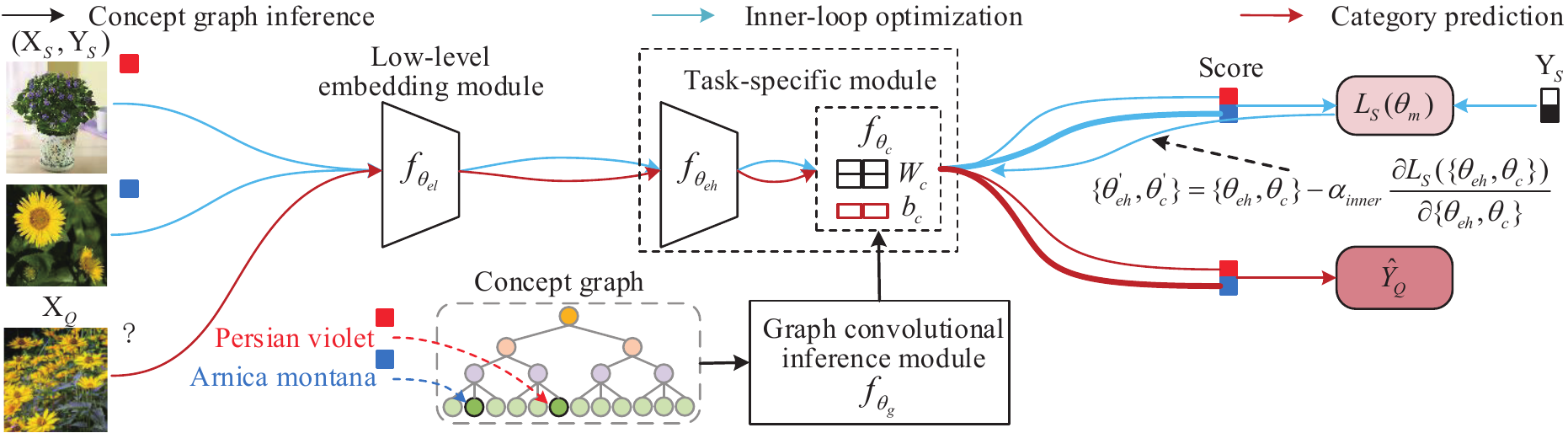}
	\end{center}
	\caption{The overview of the proposed MCIN. Here, we take a low-dimension feature embedding module as an example, which produces feature embeddings of dimension 2 for each image. For each task: (1) the phase of concept graph inference predicts a initial classifier via the selected classes node on concept graph; (2) the phase of inner-loop optimization finetunes the task-specific module by given support set; and (3) the phase of category prediction predicts the labels for query images.}
	\label{fig:framework}
\end{figure*}

Meta concept inference network (MCIN) is a cross-modal and universal meta-learner for few-shot entity and concept classification tasks, aiming to model the probability $P(y|x, S, G, \theta )$ defined in Eq. (\ref{con:3_2_1}) for task $\tau$. The MCIN framework is illustrated in Figure~\ref{fig:framework}, which consists of a low-level feature embedding module $f_{\theta_{el}}()$, a task-specific module including a high-level feature embedding module $f_{\theta_{eh}}()$ and a softmax-based classifier $f_{\theta_c}()$, and a graph convolutional inference module $f_{\theta_g}()$(for clarity, the module will be disscussed in Section~\ref{GCIM}). Here, the $\theta_{el}$, $\theta_{eh}$, and $\theta_g$ denote the optimizable parameters, where $\theta = \{\theta_{el}, \theta_{eh}, \theta_g\}$. Specifically, different from the existing meta-learning methods, we divide the feature embedding module of MCIN into two submodules: low-level feature embedding module $f_{\theta_{el}}()$ and high-level feature embedding module $f_{\theta_{eh}}()$. By doing so, the task-specific features at different abstract levels can be extracted. Here, 1) the module $f_{\theta_{el}}()$ is shared by all few-shot classification tasks, which accounts for extracting transfered features such as corners, color, and textures \cite{zoubin2014how}; 2) the module $f_{\theta_{eh}}()$ is a meta-learning module, which can quickly generate a task-specific embedding module $f_{\theta_{eh}'}()$ for a novel task and extract task-specific object features at different abstract levels. In addition, the softmax-based classifier $f_{\theta_c}()$ is also a task-specific module. However, the parameter $\theta_c$ cannot be meta-learned but infered by the module $f_{\theta_g}()$. The module $f_{\theta_g}()$ is a graph-based inference module, which is leveraged to infer the initial parameter $\theta_c$ for the task-specific classifier $f_{\theta_c'}()$ by the concept graph. 

Formally, we take the parameters $\theta_{c}$, $\theta_{c}'$, and $\theta_{eh}'$ as hidden variables, instead of optimizable parameters. For the task $\tau$, the probability $P(y|x, S, G, \theta_{el}, \theta_{eh}, \theta_g)$ of the output $y$ can be calculated by determining three conditional probabilities, namely, the conditional probability $P(\theta_c|G, \theta_g)$ of the hidden variable $\theta_{c}$, the conditional probability $P(\{\theta_{eh}', \theta_c'\}|\theta_c, S, \theta_{el}, \theta_{eh})$ of the hidden variable set $\{\theta_{eh}', \theta_{c}'\}$, and the conditional probability $P(y|x, \theta_{el}, \theta_{eh}', \theta_c')$ of the output $y$. According to the Bayesian theory, we infer the probability $P(y|x,S,G,\theta_{el}, \theta_{eh}, \theta_g)$ by leveraging the three conditional probabilities. That is,
\begin{equation}
\begin{aligned}
P(y|x, S, G,\theta_{el}, \theta_{eh}, \theta_g) = \int_{\{\theta_{eh}', \theta_c'\}}\int_{\theta_c}P(y|x, \theta_{el}, \theta_{eh}', \theta_c') P(\{\theta_{eh}', \theta_c'\}|\theta_c, S, \theta_{el}, \theta_{eh})P(\theta_c|G, \theta_g) d\theta_c d\{\theta_{eh}', \theta_c'\}
\end{aligned}
\label{con:3_3_2}
\end{equation}
where $P(\theta_c|G, \theta_g)$ and $P(\{\theta_{eh}', \theta_c'\}|\theta_c, S, \theta_{el}, \theta_{eh})$ are expressed in terms of delta function in the MCIN framwork. Specifically, the three conditional probabilities can be calculated via the following three inference steps.

\begin{itemize}[]
	\setlength{\itemsep}{0pt}
	\setlength{\parsep}{0pt}
	\setlength{\parskip}{0pt}
	\item {\bf Concept graph inference.} The step makes use of the module $f_{\theta_g}()$ to predict the initial parameter $\theta_c$ by aggregating the abstract concepts and extracting discriminated relationship among classes on the concept graph $G$ for the task-specific classifier $f_{\theta_c'}()$. It aims to transfer the abstract concepts from semantic space to vision classifier, as shown in the black lines of Figure~\ref{fig:framework}. That is, 
	\begin{equation}
	\theta_c = f_{\theta_g}(G)
	\label{con:3_3_3}
	\end{equation}
	\item {\bf Inner-loop optimization.} The hidden variable set $\{\theta_{eh}', \theta_c'\}$ is determined by applying $k$-step gradient descent on the support set $S$ of the task $\tau$, which aims to fine-tune the initial task-specific module by a few annotated samples, as shown in the blue lines of Figure~\ref{fig:framework}. For example, when we apply one step of gradient descent, the parameter set $\{\theta_{eh}', \theta_c'\}$ can be expressed as: 
	\begin{equation}
	\{\theta_{eh}', \theta_c'\} = \{\theta_{eh}, \theta_c\} - \alpha_{inner} \frac{\partial L_{(x,y) \in S}(\{\theta_{eh}, \theta_c\})}{\partial \{\theta_{eh}, \theta_c\}} 
	\label{con:3_3_4}
	\end{equation}
	where $L()$ denotes a cross-entropy loss function and $\alpha_{inner}$ is the learning rate of inner-loop optimization. 
	\item {\bf Category prediction.} The probability estimation $\hat{y}$ of each class can be found by applying the task-specific module $f_{\theta_{eh}'}()$ and $f_{\theta_c'}()$ on the feature embedding $f_{\theta_{el}}(x)$ of query samples $x$, as shown in the red lines of Figure~\ref{fig:framework}. That is,
	\begin{equation}
	\hat{y} = softmax(W_c'^{\mathrm{T}}f_{\theta_{eh}'}(f_{\theta_{el}}(x))+b_c')
	\label{con:3_3_5}
	\end{equation}
	where $W_c'$ and $b_c'$ are acquired from the hidden variable $\theta_c'$.
\end{itemize}

\begin{figure*}
	\begin{center}
		%\fbox{\rule{0pt}{2in} \rule{.9\linewidth}{0pt}}
		\includegraphics[width=1.0\linewidth]{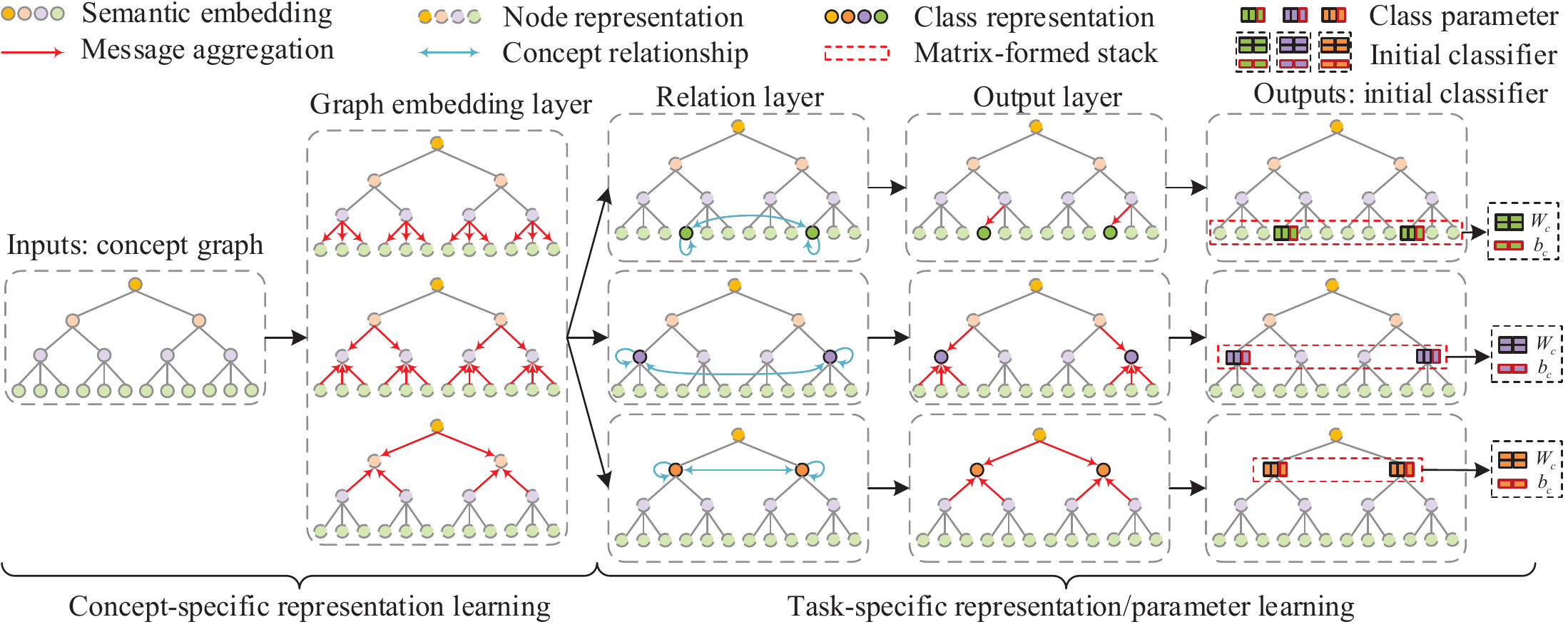}
	\end{center}
	\caption{An example of GCIM inference on a concept graph with three abstract levels, where the GCIM takes the concept graph as inputs and outputs three task-specific initial classifiers for 2-way few-shot classification tasks.}
	\label{fig:GCMN}
\end{figure*}

%-------------------------------------------------------------------------
\subsection{Graph Convolutional Inference Module}
\label{GCIM}

The graph convolutional inference module (GCIM) $f_{\theta_g}()$ introduced in Section~\ref{MCIN} can be implemented by using a novel multi-hop GCN. As illustrated in Figure~\ref{fig:GCMN}, the GCIM consists of a graph embedding layer $f_{\theta_{g_e}}()$, a relation layer $f_{\theta_{g_r}}()$, and an output layer $f_{\theta_{g_o}}()$, where $\theta_{g_e}$,  $\theta_{g_r}$, and  $\theta_{g_o}$ denote the optimizable parameters such that $\theta_{g}= \{\theta_{g_e}, \theta_{g_r}, \theta_{g_o}\}$.

The inference are carried out on three steps:

\begin{itemize}[]
	\setlength{\itemsep}{0pt}
	\setlength{\parsep}{0pt}
	\setlength{\parskip}{0pt}
	\item {\bf Step 1.}The concept graph $G$ is fed through the graph embedding layer $f_{\theta_{g_e}}()$, aiming to produce the concept-specific representation $Z_{v_i}$ for each node $v_i$:
	\begin{equation}
	Z_{v_i} = f_{\theta_{g_e}}(G), i=0,1,...,M-1
	\label{con:3_4_1}
	\end{equation}
	We employ a graph convolution with a simple layer-wise propagation rule to implement $f_{\theta_{g_e}}()$ so as to learn the abstract relationship among concepts/entities. That is,
	\begin{equation}
	Z^{h+1} = \sigma(D^{-1}AZ^{h}W_{ge}^h+b_{ge}^h)
	\label{con:3_4_2}
	\end{equation}
	where $W_{ge}^h$ and $b_{ge}^h$ are layer-specific optimizable parameters, such that $\theta_{g_e}=\left\{ W_{ge}^h, b_{ge}^h \right\}_{h=0}^{N_{ge}-1}$, $N_{ge}$ is the hop number, $Z^0$ denotes the concept semantic embedding, $A$ denotes the adjacency matrix, $D$ denotes the degree matrix, and $\sigma()$ is the activation function.
	\item {\bf Step 2.} We employ a multi-layer perceptron (MLP) to implement the relation layer $f_{\theta_{g_r}}()$, aiming to further learn a task-specific class representation $Z_{C_i}$ with discriminated relationship among classes for task $\tau$. Specifically, the representation of each class pair $C_i$ and $C_j$ of task $\tau$ are combined by the feature concatenation $cat(Z_{C_i}, Z_{C_j})$. These combined features are then fed through the relation layer to produce a representation $R_{C_i, C_j}$ for the relationship between $C_i$ and $C_j$, known as concept relationship. That is,
	\begin{equation}
	R_{C_i, C_j} = f_{\theta_{g_r}}(Z_{C_i}, Z_{C_j}) = MLP(cat(Z_{C_i}, Z_{C_j}))
	\label{con:3_4_3}
	\end{equation}
	where $i, j = 0,1,...,N-1$. Finally, the task-specific class representation $Z_{C_i}$ with the concept relationship is produced by element-wise mean over the embedding of all classes and adding the residual connection to oneself for each class, as shown in Eq. (\ref{con:3_4_4}).
	\begin{equation}
	Z_{C_i} = Z_{C_i} + \frac{1}{N}\sum_{j=0}^{N-1} R_{C_i, C_j}
	\label{con:3_4_4}
	\end{equation}
	\item {\bf Step 3.} We employ a single-layer graph convolution with normlization to model the output layer $f_{\theta_{g_o}}()$ so as to predict the initial parameter $\theta_{c}$ for the task-specific classifier. The initial parameter $\theta_{c}$ is constructed by stacking the produced task-specific class parameter of the selected classes $\left\{C_i\right\}_{i=0}^{N-1}$ for the task $\tau$. Specifically, $\theta_{c}$ can be expressed as:
	\begin{equation}
	\theta_{c} = Stack(Normalize(D^{-1}AZW_{go}+b_{go})\beta, \left\{C_i\right\}_{i=0}^{N-1})
	\label{con:3_4_5}
	\end{equation}
	where $\theta_{g_o} = \{W_{go}, b_{go}\}$, $Normalize()$ is a normalization function, $\beta$ is a super-parameter adjusting the normalization scale, and $Stack()$ is a matrix-formed stack operation for the weight $W_c$ and bias $b_c$ of the initial classifier $f_{\theta_c}()$.
\end{itemize}

%-------------------------------------------------------------------------
\subsection{Training}
The aim of meta-training is to learn to abstract for various entities via the concept graph defined by Eq. (\ref{con:3_2_1}). Therefore, the meta-objective of MetaConcept can be expressed as:
%\begin{equation}
%\min\limits_{\theta_i, \theta_g} L_{x,y \sim Q}(\theta_c') + \lambda L_{x,y \sim %Q}(\theta_c')}
%\end{equation}
\begin{equation}
\begin{aligned}
\min\limits_{\theta_{el}, \theta_{eh}, \theta_g} \lambda_e \mathbb{E}_{\tau \sim T^{e}} \left[\mathbb{E}_{S,Q \sim \tau} L_{(x,y) \sim Q}(\{\theta_{eh}', \theta_c'\})\right] + \lambda_c \sum_{l=0}^{N^{le}-1}\mathbb{E}_{\tau \sim T_l^c} \left[\mathbb{E}_{S,Q \sim \tau} L_{(x,y) \sim Q}(\{\theta_{eh}', \theta_c'\})\right]
\end{aligned}
\label{con:3_5_1}
\end{equation}
where the hidden variable set $\{\theta_{eh}', \theta_c'\}$ is performed in Eq. (\ref{con:3_3_3}) and Eq. (\ref{con:3_3_4}). We update all parameters by stochastic gradient descent optimizer (SGD) under the learning rate $\alpha_{outer}$, aiming to minimize the loss as defined in Eq. (\ref{con:3_5_1}) by applying the episode-based training strategy \cite{vinyals2016matching}, known as outer-loop optimization. 

\subsection{Overall Implementation}

The algorithm is summarized in Algorithm 1. For each episode, we sample batches of few-shot entity classification tasks from the train set. The parameters of the task-specific module are determined by infering initial classifier in Line 3 and fine-tuning the initial task-specific module in Line 4 for each task (Lines 2-5). And then we sample batches of few-shot concept classification tasks from weakly-labeled dataset and perform the inference of task-specific module at each level (Lines 6-9). Finaly, we calculate the loss with the MLCA regularization, and optimize the parameters $\theta_i$ and $\theta_g$ by using the SGD optimizer as shown in Lines 10-11.

\renewcommand{\algorithmicrequire}{\textbf{Input:}}
\renewcommand{\algorithmicensure}{\textbf{Iteration:}}
\newcommand\LONGCOMMENT[1]{%
	\hfill\#\ \begin{minipage}[t]{\eqboxwidth{COMMENT}}#1\strut\end{minipage}%
}
\begin{algorithm}[htbp]
	\caption{Meta-training procedure of MetaConcept}
	\begin{algorithmic}[1]
		\REQUIRE ~~\\ 
		Concept graph $ G=(V,E)$ with $M$ nodes, $N^l$ abstract levels, and $d$-dimension concept semantic matrix $Z\in \mathbb{R}^{M \times d}$; train set $D^{tr}$; learning rates of inner loop and outer loop $\alpha_{inner}$, $\alpha_{outer}$; super-parameters $\beta$ and $\lambda$.
		\renewcommand{\algorithmicrequire}{\textbf{Output:}}
		\REQUIRE ~~\\ 
		Low-level and high-level feature embedding module $f_{\theta_{el}}()$, $f_{\theta_{eh}}()$ and graph convolutional inference module $f_{\theta_g}()$.
		\renewcommand{\algorithmicrequire}{\textbf{Initialization:}}
		\REQUIRE ~~\\ 
		Randomly initialize model parameters $\theta_{el}$, $\theta_{eh}$ and $\theta_g$.
		\ENSURE ~~\\
		\FOR {each episode}
		% sample multi-levels few-shot tasks
		\STATE ($S_l$, $Q_l$) = RandomSample($D^{tr}$);
		\STATE Predict initial parameter for task-specific classifier by the concept graph $G$ in Eq. (\ref{con:3_4_1})- Eq. (\ref{con:3_4_5});
		\STATE Apply $k$-step gradient descent for initial task-specific module on support set $S_l$ in Eq. (\ref{con:3_3_4});
		\STATE Calculate valid loss of query set $Q_l$;
		\FOR {each level $l$ in $G$}
		\STATE ($S_l$, $Q_l$) = RandomSample($D_l^{we}$);
		\STATE Perform Line 3 $\sim$ Line 5 again;  
		\ENDFOR 
		\STATE Calculate loss with MLCA regularization in Eq. (\ref{con:3_5_1});
		\STATE Optimize $\theta_{el}$, $\theta_{eh}$, and $\theta_g$ by using SGD;
		\ENDFOR
	\end{algorithmic}
\end{algorithm}
%------------------------------------------------------------------------

\section{Performance Evaluation}

In this section, we first discuss the experiment results on two setting of WSFSL defined in Section \ref{section:3_2}, followed by our ablation study.
\subsection{Datasets and Settings}

{\bf WS-ImageNet-Pure.} The dataset \cite{liu2019prototype} is a subset of 188 classes selected from the ILSVRC-12 dataset at five different levels (level-7, level-6, level-5, level-4, level-3) of the ImageNet WordNet hierarchy, where the classes from level-7 are the concrete entity classes and the classes from other four levels are the abstract concept classes. The dataset is splited into two disjoint subsets following \cite{liu2019prototype}, i.e.\ a meta-training set and a meta-test set. Note that the data samples of all classes are sampled in a bottom-up manner, where the samples of any classes on level $l$ are sampled from all the images belonging to the class in ImageNet. Further details can be found in \cite{liu2019prototype}

{\bf WS-ImageNet-Mix.} The dataset \cite{liu2019prototype} is another subset of 188 classes selected from the ILSVRC-12 dataset. We still adopt the same split method in \cite{liu2019prototype}. The dataset are similiar to WS-ImageNet-Pure. The key difference is that the data samples of the abstract concept class can belong to the remaining $20\%$ level-7 classes outside of the $80\%$ level-7 classes used for generating few-shot entity classification tasks. The goal is to further analyze the effect of the abstract concept classes when its data samples sampled from other concrete entities not involved in the few-shot entity classification tasks. Please refer to \cite{liu2019prototype} for details.

{\bf Concept Graph.} The concept graph can be constructed from cheap knowledge graph such as WordNet \cite{miller1995wordnet} for each given dataset. Specifically, we regard all categories of the dataset as the leaf nodes, and then extract their abstract concepts from knowledge graph at multiple levels as the nonleaf nodes to build the concept graph. We use the GloVe model \cite{pennington2014glove} to extract the 300-dimension concept semantic embedding for the concrete entities and the abstract concepts, that the mean value of word embeddings of entity and concept names. 

\subsection{Implementation Details}
\label{section:4_2}

{\bf Architecture. } We use a 4-layer convnet \cite{liu2019prototype, snell2017prototypical, liu2019learning} with 64 channels per layer for a fair comparison, which is partitioned into a low-level module with two layers and a high-level module with two layers. In the meta concept inference network, we use two-layer graph convolution to model the graph embedding layer whose dimensions are 4096 and 2048 respectively, where we add dropout layers \cite{krizhevsky2012imagenet} with keep probability of 0.9. Moreover, we use a two-layer MLP to model the relation layer whose dimensions are 4096 and 2048 respectively, where we take Leaky ReLU with the negative slope of 0.1 as the activation function and add dropout layers \cite{krizhevsky2012imagenet} with keep probability of 0.9. Furthermore, we use a single-layer graph convolution as the output layer with 1601 dimensions.

{\bf Training details. }We adopt the SGD optimizer with an initial learning rate of $10^{-1}$, a momentum of 0.9, and weight decay of 0.0005 to train the proposed model with 20000 iterations, where the learning rate is reduced by 0.1 for every 5000 iterations. Hyperparameters $\beta$ and $\lambda_c$, and inner-learning rate $\alpha_{inner}$ are set to be 0.2, 1.0, and 0.01, respectively. For the inner update step $k$, we set to be 5 in the meta-training phase and 20 in the meta-test phase.  

{\bf Experimental setup}
All experiments are evaluated by 5-way 1-shot and 5-way 5-shot classification task on the above datasets. We report the average classification accuracies with the $95\%$ confidence intervals obtained by sampling 600 episodes from the meta-test set. For comparison, we implement MLCA by using the abstract concepts at level-6, level-5, level-4, and level-3.

\subsection{Discussion of Results}
In this section, we conduct two experimental setting of WSFSL defined in Section \ref{section:3_2} on the public two WFSFL datasets (i.e.\ WS-ImageNet-Pure and WS-ImageNet-Mix) to show the effectiveness of proposed MetaConcept.

\subsubsection{Results on the default setting of WSFSL ($\lambda_e=1.0$)}
\label{section_4_3_1}
We compare MetaConcept method with the recent state-of-the-art methods on the above datasets, and show the results of few-shot entity classification tasks with three different aspects.
\begin{itemize}[]
	\setlength{\itemsep}{0pt}
	\setlength{\parsep}{0pt}
	\setlength{\parskip}{0pt}
	\item We reimplement a number of classical and latest methods without exploring concept hierarchy knowledge as the basis of comparison, and report the mean accuracies as the first set of results in Table~\ref{table:result}, where we adopt the 4-layer convnet as feature extractor. Note that these methods only are trained on finely-labeled data (i.e.\ the concrete entity level), ignoring the weakly-labeled data. The goal is to show the effectiveness of exploring concept hierarchy by using weakly-labeled data and concept graph prior information.
	\item We present results in \cite{liu2019prototype} for comparison as the second set of results in Table~\ref{table:result}. Different from MetaConcept, the methods explore the weakly-labeled data by propagating prototypes on the subgraph structure of concept graph. The goal is to show the effectiveness of the proposed MetaConcept method on WSFSL problem.
	\item We report the results of MetaConcept as shown in the last row in Table~\ref{table:result}.
\end{itemize}

The results are presented in 4-tupled values. Here, the two sets of numbers in a tuple corrspond to the experimental results using WS-ImageNet-Pure and WS-ImageNet-Mix, respectively. The two numbers on each set of numbers on a tuple correspond to the mean classification accuracy averaged over 600 test episodes on the 5-way 1-shot and 5-way 5-shot tasks, respectively. The 95\% confidence intervals of the estimates are also shown in Table~\ref{table:result}. The classification accuracy is defined as the number of correct predictions divided by the total number of query samples in an episode. We find that our proposed MetaConcept achieves the best few-shot classification performance of ($50.29\%$, $58.02\%$, $38.02\%$, $47.39\%$, $59.38\%$, $34.48\%$) and achieves a significant improvement ranging from 2\% to 6\% on the above six tasks. This verifies the effectiveness of our proposed MetaConcept.

{\bf Performance analysis of exploring concept hierarchy.} The comparison results of first set and last row of Table~\ref{table:result} exhibit the impact of exploring concept hierarchy knowledge by the leveraging the concept graph and weakly-labeled data on classification performance. We can see that MetaConcept which explores concept hierarchy knowledge outperforms the baseline methods without this, around 3\% to 12\% on all tasks. Moreover, it is obvious that the improvement on the 5-way 1-shot tasks is larger than that on the 5-way 5-shot tasks, i.e.\ around $4.18\%$ and $1.38\%$ on WS-ImageNet-Pure and WS-ImageNet-Mix, respectively. This verifies the effectiveness of exploring concept hierarchy knowledge by leveraging the concept graph and weakly-labeled data and implies that it can significantly boost the performance on novel tasks especially when the annotated samples are insufficient. 

\begin{table*}
	\caption{Experiment results on WS-ImageNet-Pure and WS-ImageNet-Mix when $\lambda_e$ is set to one. The best results of each set are highlighted in bold and the best result are italized. Note that the results from the results reported in \cite{liu2019prototype} are denoted with $^*$ and WS denotes the weakly-supervised strategy on subgraph structure.}
	\label{table:result}
	\begin{tabular}{lcccc}
		\toprule
		\multicolumn{1}{l}{\multirow{2}{*}{Method}}& \multicolumn{2}{c}{WS-ImageNet-Pure} & \multicolumn{2}{c}{WS-ImageNet-Mix} \\ 
		\cline{2-5}
		& 5-way 1-shot & 5-way 5-shot & 5-way 1-shot & 5-way 5-shot \\
		\midrule
		MAML \cite{finn2017model} & $30.19 \pm 0.86\%$  & $46.45 \pm 1.00\%$ & $34.23 \pm 0.89\%$  & $40.45 \pm 0.99\%$ \\
		ProtoNet $^*$ \cite{snell2017prototypical} & $33.17 \pm 1.65\%$  & $46.76 \pm 0.98\%$ & $31.93 \pm 1.62\%$  & $49.80 \pm 0.90\%$ \\
		GNN $^*$ \cite{satorras2018few} &  $30.83 \pm 0.66\%$  & $41.33 \pm 0.62\%$ & $33.60 \pm 0.11\%$  & $45.87 \pm 0.12\%$ \\
		CloserLook $^*$ \cite{chen2019closer} &  $32.27 \pm 1.58\%$  & $46.02 \pm 0.74\%$ & $33.10 \pm 1.57\%$  & $40.67 \pm 0.73\%$ \\
		MetaOptNet-SVM \cite{lee2019meta} & $37.85 \pm 0.97\%$  & $49.17 \pm 0.98\%$ & $40.91 \pm 0.92\%$  & $54.38 \pm 0.95\%$ \\
		LEO \cite{rusu2018meta} &  $37.38 \pm 0.78\%$  & $48.83 \pm 0.73\%$ & $37.92 \pm 0.72\%$  & $49.19 \pm 0.71\%$ \\
		FSLKT \cite{zhimao2019few} & 39.56 $\pm$ 0.86$\%$ & 45.48 $\pm$ 0.95$\%$ & 39.98 $\pm$ 0.87$\%$ & 48.81 $\pm$ 0.97$\%$ \\
		AM3-ProtoNet \cite{xing2019adaptive} &  $36.82 \pm 0.92\%$  & $49.41 \pm 1.01\%$ & $38.58 \pm 0.96\%$  & $54.99 \pm 0.96\%$ \\
		TPN \cite{liu2019learning} & \textbf{39.95} $\pm$ \textbf{0.79}$\%$ & \textbf{51.86} $\pm$ \textbf{0.78}$\%$ & \textbf{42.04} $\pm$ \textbf{0.84}$\%$ & \textbf{55.41} $\pm$ \textbf{0.70}$\%$ \\
		\hline
		WS-ProtoNet $^*$ \cite{snell2017prototypical} & $32.13 \pm 1.48\%$ & $44.41 \pm 0.93\%$ & $31.80 \pm 1.48\%$ & $49.03 \pm 0.93\%$\\
		WS-GNN(2018)$^*$\cite{satorras2018few} & $32.33 \pm 0.52\%$ & $45.67 \pm 0.87\%$ & $30.33 \pm 0.80\%$ & $47.33 \pm 0.28\%$\\
		WS-CloserLook(2019)$^*$\cite{chen2019closer} & $32.63 \pm 1.55\%$ & $43.76 \pm 0.93\%$ &$31.13 \pm 1.51\%$ & $44.90 \pm 0.78\%$ \\
		PPN $^*$ \cite{liu2019prototype} & $37.37 \pm 1.64\%$ & $50.31 \pm 1.00\%$& $36.23 \pm 1.69\%$ & $52.38 \pm 0.92\%$ \\
		PPN+ $^*$ \cite{liu2019prototype} & \textbf{48.00} $\pm$ \textbf{1.70}$\%$ & \textbf{52.36} $\pm$ \textbf{1.02}$\%$ & \textbf{41.60} $\pm$ \textbf{1.67}$\%$ & \textbf{53.95} $\pm$ \textbf{0.96}$\%$\\
		\hline
		MetaConcept & \emph{\textbf{50.29}} $\pm$ \emph{\textbf{0.90}}$\%$ & \emph{\textbf{58.02}} $\pm$ \emph{\textbf{0.93}}$\%$ & 
		\emph{\textbf{47.39}} $\pm$ \emph{\textbf{0.96}}$\%$ & \emph{\textbf{59.38}} $\pm$ \emph{\textbf{0.98}}$\%$\\
		\bottomrule
	\end{tabular}
\end{table*}

\begin{table*}
	\caption{Experimental results on WS-ImageNet-Pure and WS-ImageNet-Mix when $\lambda_e$ is set to zero. The best results of each set are highlighted in bold and the best results are italized. Note that ``MLCA-'' denotes combining with MLCA technique.}
	\label{table:result_ws}
	\begin{tabular}{lcccccc}
		\toprule
		\multicolumn{1}{l}{\multirow{2}{*}{Method}}& \multicolumn{2}{c}{WS-ImageNet-Pure} & \multicolumn{2}{c}{WS-ImageNet-Mix} \\ 
		\cline{2-5}
		& 5-way 1-shot & 5-way 5-shot & 5-way 1-shot & 5-way 5-shot \\
		\midrule
		MLCA-MAML & $33.25 \pm 0.91\%$  & $45.49 \pm 1.02\%$ & $35.60 \pm 0.87\%$  & $48.67 \pm 0.93\%$ \\
		MLCA-ProtoNet & $34.60 \pm 0.92\%$  & $47.89 \pm 1.02\%$ & $35.16 \pm 0.96\%$  & $52.65 \pm 0.97\%$ \\
		MLCA-MetaOptNet-SVM & $35.76 \pm 0.88\%$  & $48.07 \pm 0.95\%$ & $39.64 \pm 0.97\%$  & $50.73 \pm 0.94\%$ \\
		MLCA-LEO &  $33.74 \pm 0.71\%$  & $47.18 \pm 0.72\%$ & $37.39 \pm 0.73\%$  & $49.84 \pm 0.74\%$ \\
		MLCA-FSLKT & \textbf{45.22} $\pm$ \textbf{0.90}$\%$ & 50.03 $\pm$ 0.93$\%$ &  \textbf{40.23} $\pm$ \textbf{0.90}$\%$ & 49.85 $\pm$ 0.95$\%$ \\
		MLCA-AM3-ProtoNet &  $40.07 \pm 0.89\%$  & \textbf{50.55} $\pm$ \textbf{0.95}$\%$ & $37.87 \pm 0.92\%$  & \textbf{53.99} $\pm$ \textbf{0.93}$\%$ \\
		MLCA-TPN & 39.20 $\pm$ 0.79$\%$ & 49.19 $\pm$ 0.76 $\%$ & 39.98 $\pm$ 0.83$\%$ & $53.61 \pm 0.73\%$ \\
		\hline
		MetaConcept & \emph{\textbf{48.56}} $\pm$ \emph{\textbf{0.93}}$\%$ &  \emph{\textbf{56.17}} $\pm$ \emph{\textbf{0.93}}$\%$ & \emph{\textbf{47.23}} $\pm$ \emph{\textbf{1.00}}$\%$ & \emph{\textbf{57.45}} $\pm$ \emph{\textbf{0.95}}$\%$ \\
		\bottomrule
	\end{tabular}
\end{table*}

{\bf Performance analysis of MetaConcept.} The comparison results of MetaConcept, WS-ProtoNet, WS-GNN, WS-CloserLook, and Prototype Propagation Networks methods are shown in the second set and last row of Table~\ref{table:result}. We can also observe that MetaConcept consistently outperforms the baseline methods with weakly-supervised strategy. Especially, compared with PPN+ method, MetaConcept are consistently yielding a higher classification accuracy, around 2\% to 6\% on all tasks (different from PPN+, MetaConcept does not use the weakly-labeled data during the meta-test phase). This implies that MetaConcept is more effective than the prototype propagated methods with weakly-supervised strategy. There are two reasons for such performance gain. First, the MetaConcept builds a MLCA regularization on a global concept graph, instead of a subgraph, which can guide the MCIN-based meta-learner to learn to abstract multi-level concepts via the global concept graph, i.e. fully exploiting the concept hierarchy. Second, the joint inference of the abstract concepts and a few annotated samples is more effective for infering task-specific classifiers than single vision inference of a few annotated samples. Furthermore, it is worth noting that the improved performances of our proposed MetaConcept on WS-ImageNet-Mix is remarkable, around 4\% to 6\% in classification accuracy. The observation indicates that the data samples of abstract concept classes from other concrete entities are particularly helpful for MetaConcept. It can provide more abundant abstract information. 

\subsubsection{Results on the more economical setting of WSFSL ($\lambda_e=0.0$)} 
We have conducted some detailed experiments on the more economical WSFSL defined in Section \ref{section:3_2} (i.e.\ $\lambda_e$ is set to zero in Eq. 2), aiming to show the effectiveness of our proposed MetaConcept trained only on weakly-labeled data sets. We compare MetaConcept method with seven baseline methods that are able to apply MLCA on WS-ImageNet-Pure and WS-ImageNet-Mix. Here, we replace probability $P(y|x,S,G,\theta)$ of Eq. (\ref{con:3_2_1}) with $P(y|x,S,\theta)$ so as to apply MLCA in the seven baseline methods. The results of few-shot entity classification tasks are shown in Table~\ref{table:result_ws}. According to the results of Tables~\ref{table:result} and \ref{table:result_ws}, we find that 1) MetaConcept outperforms seven baseline methods, around 3\% to 8\% in classification accuracy on all tasks; 2) MetaConcept also outperforms the PPN method trained on finely-labeled and weakly-labeled data, around 5\% to 11\% on all tasks, although it is trained only on the weakly-labeled data; and 3) for the WS-ImageNet-Mix, MetaConcept trained only on the weakly-labeled data also achieves almost consistent performance with that on the finely-labeled and weakly-labeled data. This further implies that our MetaConcept is effective which can explore and exploit concept hierarchy knowledge via concept graph and weakly-labeled data for FSL.

\subsection{Ablation Study}

In the section, we carry out an ablation study on the default setting of WSFSL ($\lambda_e=1.0$) to answer the following research questions: 1) How does MLCA affect the performance of few-shot entity classification?  2) How does MLCA affect the performance of few-shot concept classification? 3) How do MCIN and concept semantics (CS) affect the performance of few-shot entity classification? 

\begin{table*}
	\caption{Effect of MLCA and concept semantic (CS). Experiment results on two datasets with adding or removing MLCA or CS.}
	\label{table:ablation}
	\begin{tabular}{lccccccccc}
		\toprule
		\multicolumn{1}{l}{\multirow{2}{*}{Method}} & \multicolumn{1}{c}{\multirow{2}{*}{MLCA}} & \multicolumn{1}{c}{\multirow{2}{*}{CS}} & \multicolumn{2}{c}{WS-ImageNet-Pure} & \multicolumn{2}{c}{WS-ImageNet-Mix} \\ 
		\cline{4-7}
		& & & 5-way 1-shot & 5-way 5-shot & 5-way 1-shot & 5-way 5-shot \\
		\midrule
		MAML & $\surd$ &  & $35.83 \pm 0.99\%$  & $47.09 \pm 0.92\%$ & $36.35 \pm 0.93\%$  & $50.39 \pm 0.95\%$ \\	
		ProtoNet & $\surd$ &  & $35.12 \pm 0.98\%$  & $49.99 \pm 1.00\%$ & $36.25 \pm 0.91\%$  & $51.79 \pm 0.97\%$ \\	
		MetaOptNet-SVM & $\surd$ &  & $38.77 \pm 0.94\%$  & $51.43 \pm 0.96\%$ & $42.67 \pm 0.99\%$  & $57.28 \pm 0.96\%$ \\
		LEO & $\surd$ &  &  $38.35 \pm 0.80\%$  & $50.36 \pm 0.73\%$ & $38.15 \pm 0.73\%$  & $51.64 \pm 0.71\%$ \\
		FSLKT& $\surd$ & $\surd$ & $47.53 \pm 0.89\%$  & $53.13 \pm 0.91\%$ & $43.67 \pm 0.94\%$  & $53.07 \pm 0.96\%$ \\
		AM3-ProtoNet& $\surd$ & $\surd$ & $37.70 \pm 0.96\%$  & $51.18 \pm 0.97\%$ & $39.09 \pm 0.96\%$  & $57.61 \pm 0.97\%$\\
		TPN& $\surd$ &  & $39.36 \pm 0.79\%$  & $51.58 \pm 0.78\%$ & $43.17 \pm 0.85\%$  & $55.09 \pm 0.71\%$ \\
		\hline
		MetaConcept & $\surd$ & $\surd$ & \emph{\textbf{50.29}} $\pm$ \emph{\textbf{0.90}}$\%$ & \emph{\textbf{58.02}} $\pm$ \emph{\textbf{0.93}}$\%$ & 
		\emph{\textbf{47.39}} $\pm$ \emph{\textbf{0.96}}$\%$ & \emph{\textbf{59.38}} $\pm$ \emph{\textbf{0.98}}$\%$ \\
		MetaConcept &   & $\surd$  & $42.99 \pm 0.83\%$  & $51.59 \pm 0.91\%$ & $42.15 \pm 0.95\%$  & $54.52 \pm 1.01\%$ \\
		MetaConcept & $\surd$  &  & $46.59 \pm 0.88\%$  & $55.15 \pm 0.90\%$ & $43.85 \pm 0.97\%$  & $56.83 \pm 0.95\%$ \\
		\bottomrule
	\end{tabular}
\end{table*}

{\bf Effects of MLCA on entity classification.} We show the results of MetaConcept with MLCA and without MLCA, and seven baseline methods that are able to apply MLCA in Table \ref{table:ablation} to analyze the performance impact of MLCA. In the Tables \ref{table:result} and \ref{table:ablation}, we find that 1) the performance of the seven baseline methods becomes better by applying MLCA, where, for example, the MetaOptNet-SVM achieves a classification accuracy improvement of $0.5\%$ to $3\%$ on all tasks; 2) the performance of ProtoNet with MLCA outperforms ProtoNet with the weakly-supervised strategy on subgraph, around 2\% to 7\% in classification accuracy on all tasks; 3) the performance of MetaConcept becomes poor when removing the MLCA, around 4\% to 8\% reduction in classification accuracy on all tasks. The observations indicate that MLCA is essential for MetaConcept and can improve the classification performance on novel tasks significantly. 

\begin{figure}
	\begin{center}
		\includegraphics[width=0.6\linewidth]{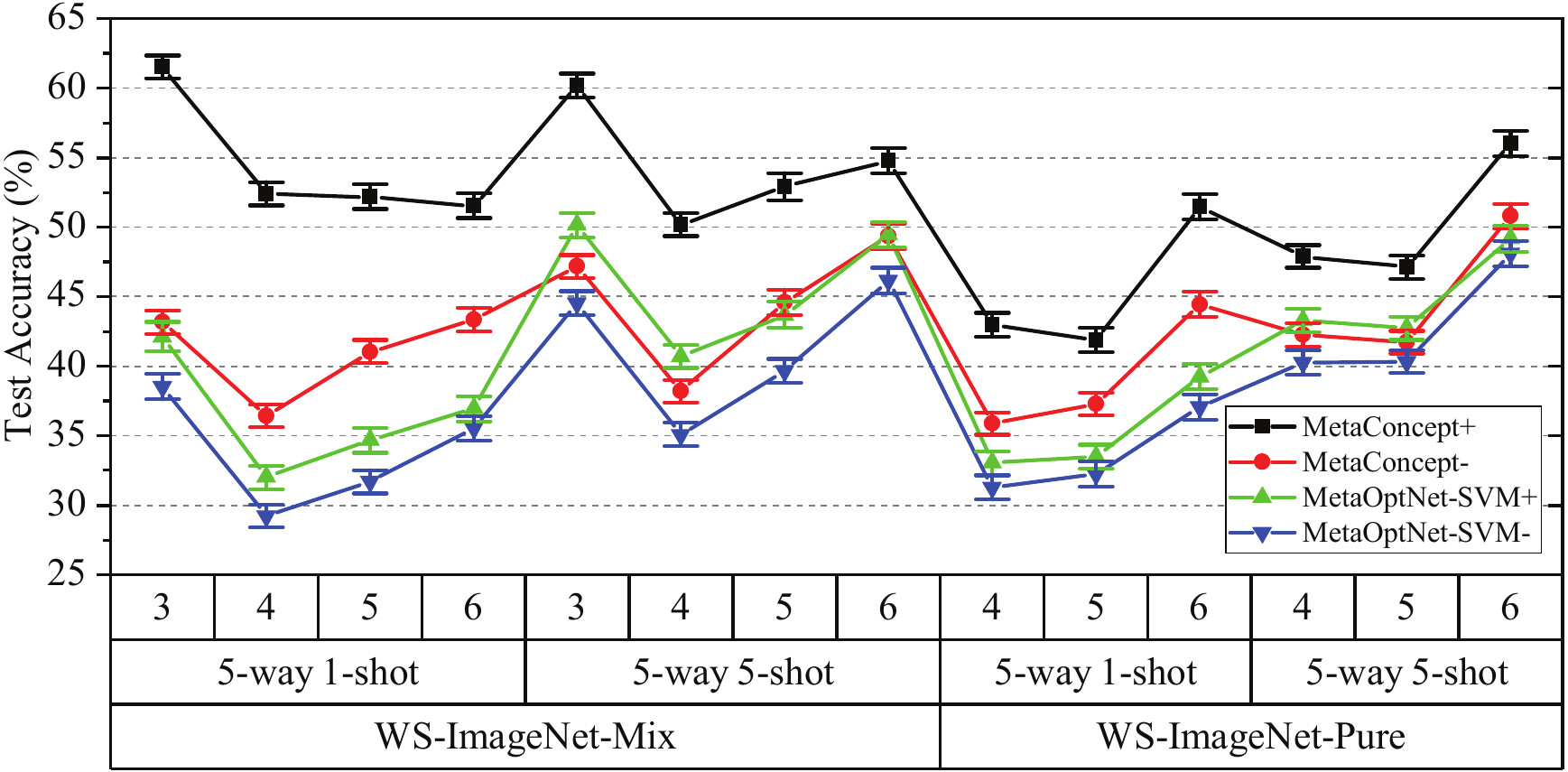}
	\end{center}
	\caption{Test accuracy of MetaConcept and MetaOptNet-SVM at abstract levels $l=3,4,5,6$ of two datasets, with MLCA (marked with $+$) and without MLCA (marked with $-$). Note that the meta-test set of WS-ImageNet-Pure has only one class at level-3.}
	\label{fig:concept_classification}
\end{figure}

{\bf Effects of MLCA on concept classification.} In Figure \ref{fig:concept_classification}, we show the results of all abstract levels by using MetaConcept and MetaOptNet-SVM so as to analyze the performance on few-shot concept classification with MLCA or without MLCA. From the results, we can see that the performance of the MetaConcept exceeds that of MetaOptNet-SVM even without MLCA. This demonstrates that the meta concept inference strategy is effective, which can be generalized to the abstract concepts from the concrete entities. On the other hand, the performance of MetaConcept can be further boosted by applying the MLCA technique. This shows the effectiveness of MLCA which can guide the meta-learner to learn to abstract concepts via the concept graph.

{\bf Effects of MCIN and concept semantics on entity classification.} In Table \ref{table:ablation}, we can also observe that the MetaConcept outperforms the seven baseline methods combined with MLCA (e.g.\ around 2\% to 12\% for AM3-ProtoNet), which testifies that the MCIN is more effective when combined with MLCA. In addition, we compare the performance of MetaConcept with and without the concept semantics. Note that we use one-hot code as the feature embedding of each node in MetaConcept without using concept semantic. This aims to infer the initial classifier by making use of graph structure of the concept graph. As shown in Table \ref{table:ablation}, we observe that MetaConcept employing the concept semantic achieves a better performance, which exceeds the MetaConcept with one-hot code, around 2\% to 5\% in classification accuracy on all tasks. Furthermore, compared with baseline methods, MetaConcept without using concept semantic still achieves better performance except the 5-way 5-shot task on WS-ImageNet-Mix. Hence, MetaConcept can also fully learn by only exploring the graph structure of concept graph. This shows that the concept semantics and MCIN employed in MetaConcept are effective for quickly adapting to a novel task.

\subsection{Hyperparameters Analysis}

\begin{figure}
	\begin{center}
		\includegraphics[width=0.6\linewidth]{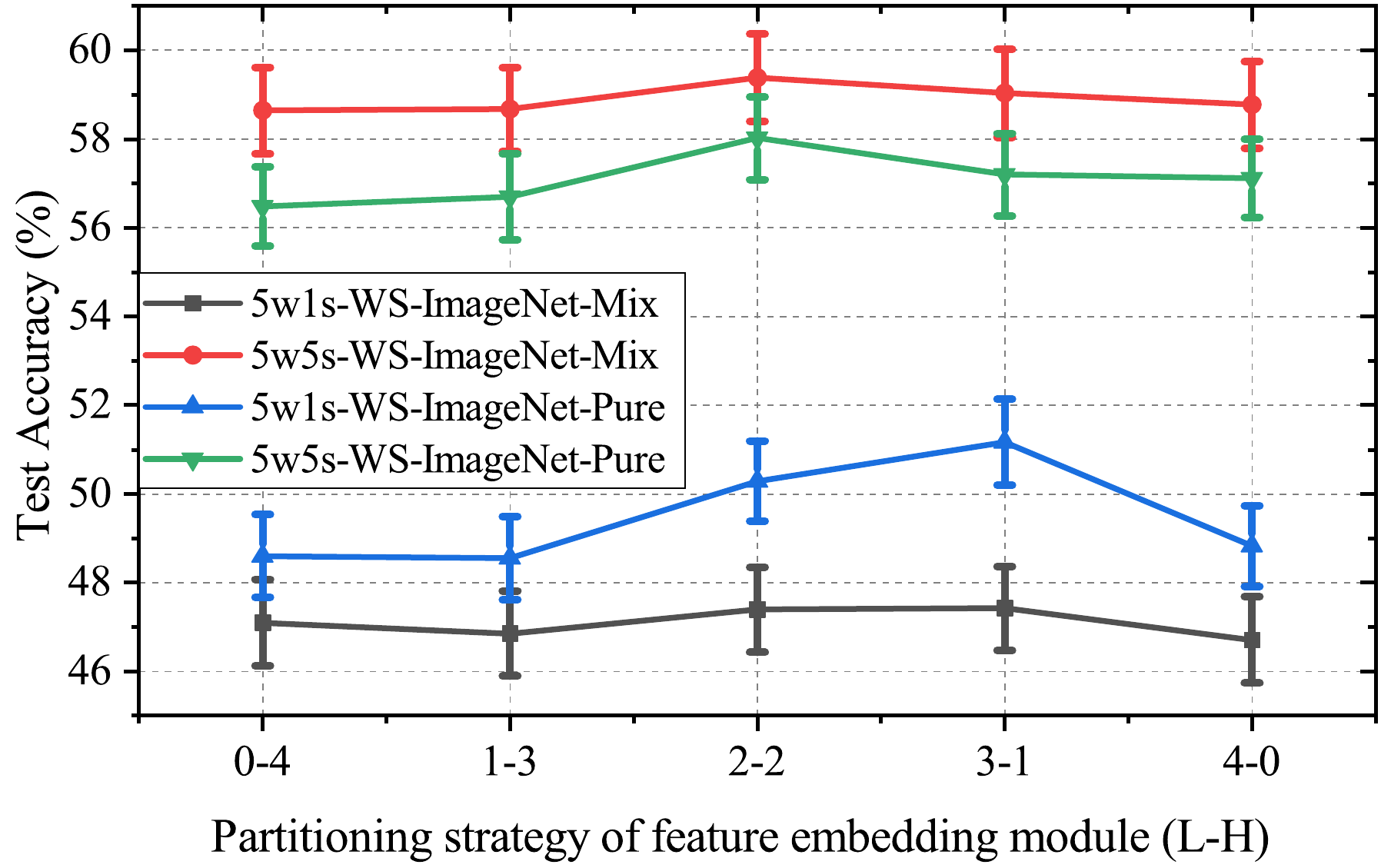}
	\end{center}
	\caption{Test accuracy on WS-ImageNet-Pure and WS-ImageNet-Mix when setting different partitioning strategies for feature embedding. Here, L and H denote the number of convolutional layer of low-level and high-level feature embedding module respectively.}
	\label{fig:partition_lines}
\end{figure}

In the section, we carry out detailed experiments on the default setting of WSFSL ($\lambda_e=1.0$) to further answer the following research questions: 1) How does the partitioning strategy of feature embedding module affect the performance of few-shot entity classification? 2) How does the weight of MLCA regularization $\lambda_c$ affect the performance of few-shot entity classification? 3) How does the normalization scale of classifier parameter $\beta$ affect the performance of few-shot entity classification? 

\begin{figure}
	\begin{center}
		\includegraphics[width=0.6\linewidth]{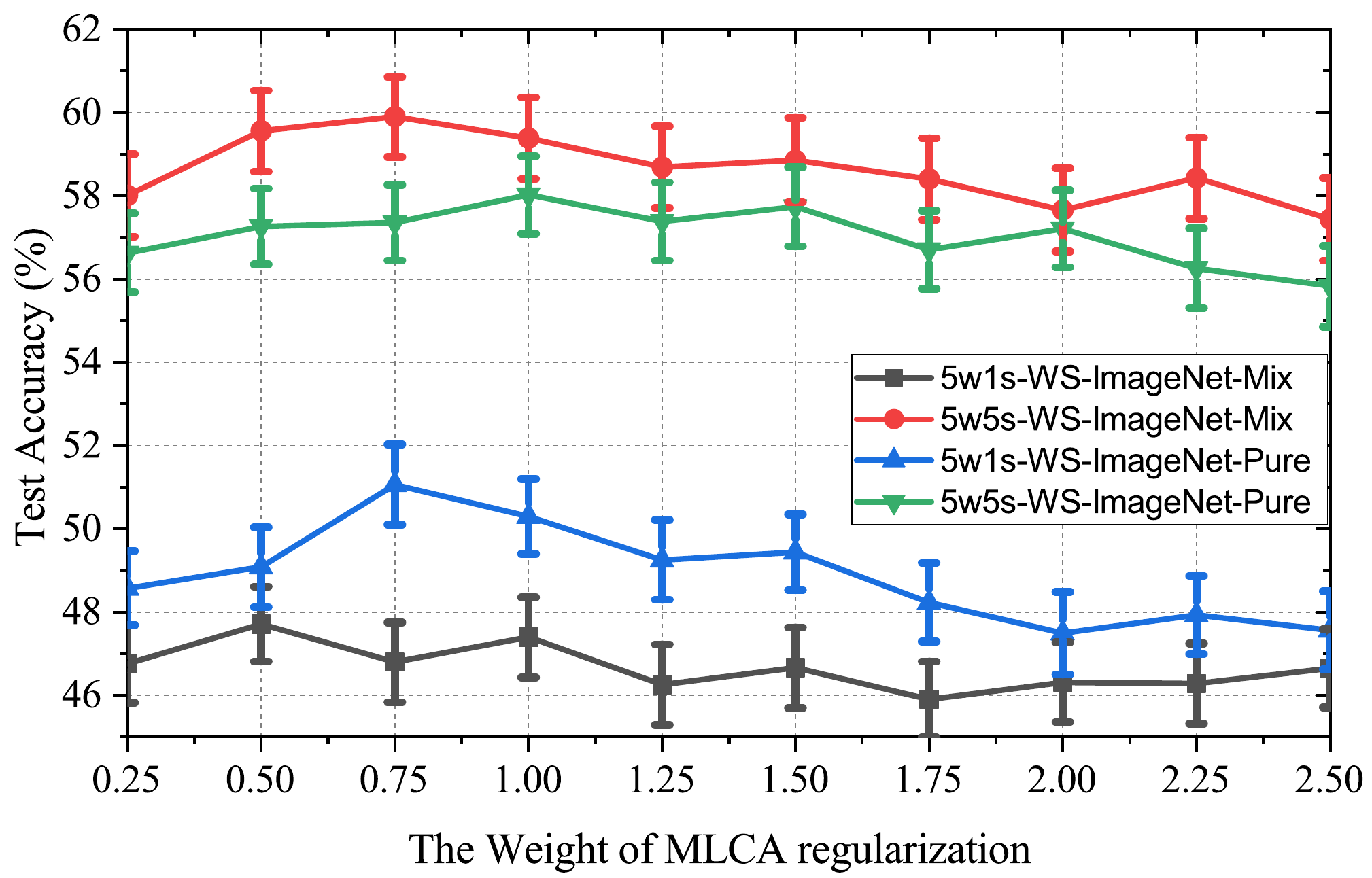}
	\end{center}
	\caption{Test accuracy on WS-ImageNet-Pure and WS-ImageNet-Mix when increasing the weight of MLCA regularization.}
	\label{fig:weight_mlca_line}
\end{figure}

\begin{figure}
	\begin{center}
		\includegraphics[width=0.6\linewidth]{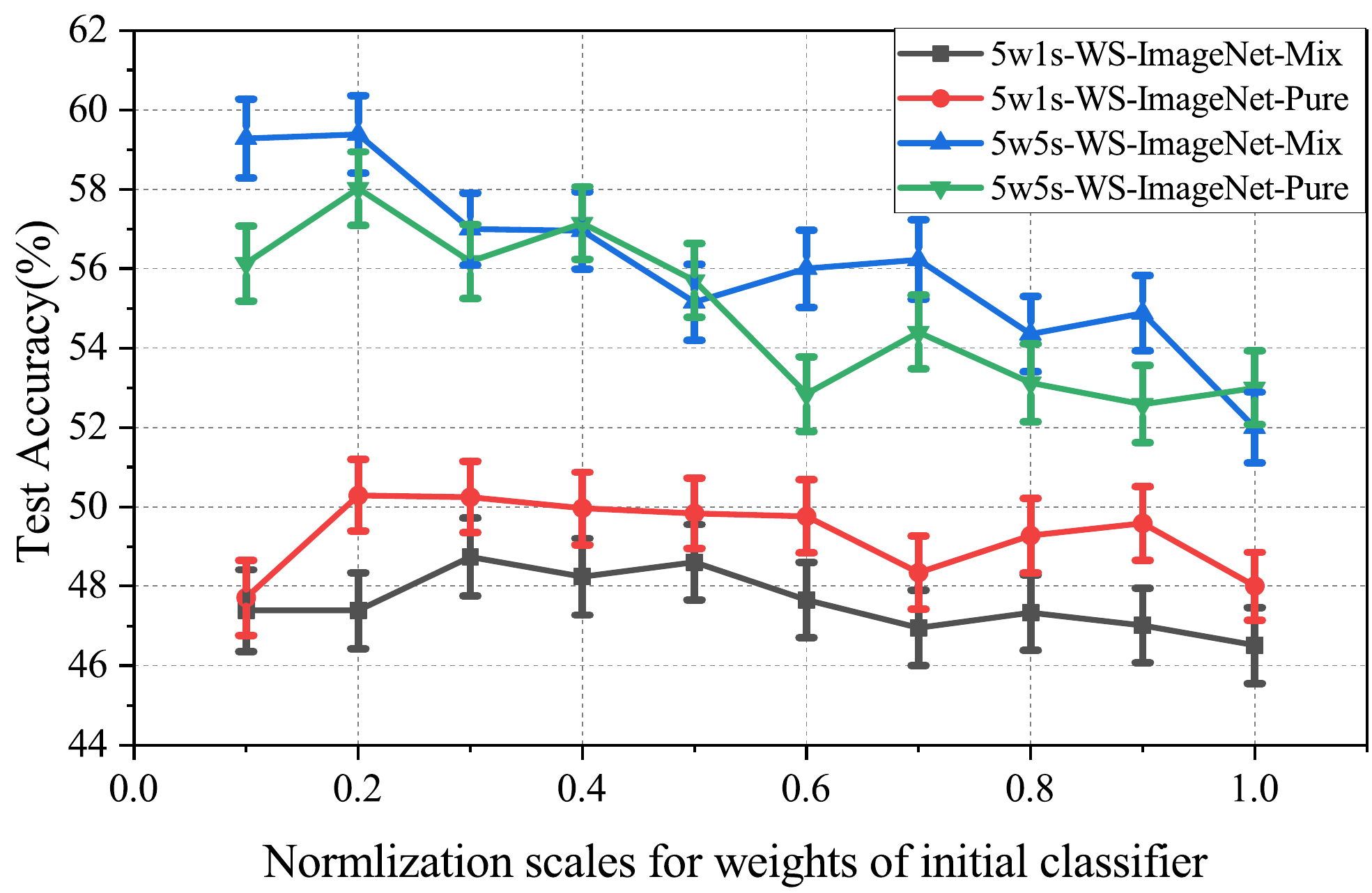}
	\end{center}
	\caption{Test accuracy on WS-ImageNet-Pure and WS-ImageNet-Mix when increasing the normization scale of classifier parameter.}
	\label{fig:normization_scale_lines}
\end{figure}

{\bf Effects of embedding module on entity classification.} We conduct a number of experiments on WS-imagenet-Pure and WS-imagenet-Mix by applying five partitioning strategies for feature embedding module, aiming to analyze the impact of the partitioning strategy on few-shot entity classification. We show the results of MetaConcept in Figure \ref{fig:partition_lines}. As shown in the  Figure \ref{fig:partition_lines}, we can find that 1) the performance of MetaConcept method with partitioning strategy (i.e.\ L-H = 1-3, 2-2, and 3-1) outperforms without partitioning strategy (i.e.\ L-H = 0-4 and 4-0); 2) the MetaConcept method achieve the best performance when partitioning the feature embedding module into low-level embedding module with two convolutional layers and high-level embedding module with two convolutional layers, i.e.\ L=2 and H=2. This shows that the partitioning strategy of embedding module is helpful for learning a cross-level meta-learner. The reason may be that 1) the low-level embedding module is a shared module, which can learn the transferable low-level feature such as corners, edge, color, and textures; 2) the high-level embedding module is a task-specific module, which can quickly adapt to a new task and extract task-specific object feature at different abstract levels.

{\bf Effects of hyperparameters on entity classification.} We show the results of MetaConcept with different normalization scale in Figure \ref{fig:normization_scale_lines} to further analyze the performance of the MetaConcept method. Here, the weight of MLCA regularization $\lambda_c$ is set to 1.0 and the normalization scale $\beta$ is varied from 0.1 to 1.0. As shown in the Figure \ref{fig:normization_scale_lines}, we can find that 1) the hyperparameter $\beta$ has important effect on few-shot entity classification task, especially 5-way 5-shot classification task; 2) the MetaConcept method can achieve better performance when we set a smaller scale for the normalization of class parameter, i.e.\ around $\beta=0.2$. In addition, we also analyze the performance of MetaConcept when using different weight of MLCA regularization in Figure \ref{fig:weight_mlca_line}. Here, the hyperparameter $\beta$ is set to 0.2 and $\lambda$ is varied from 0.25 to 2.50. It can be seen that the performance keeps increasing when we increase the weight of regularization from 0.0 to 1.0, and then decreasing, i.e.\ we can achieve best performance when seting the weight to $0.5 \sim 1.0$. 

%------------------------------------------------------------------------
\section{Conclusions}

In this paper, we shows that our proposed meta-learning method based on concept graph achieves the state-of-the-art performance for tackling weakly-supervised few-shot learning problems. We propose a novel regularzation with multi-level conceptual abstraction to model a conceptual abstract strategy, which is shown to be effective for minimizing the generalization error of base learner across tasks via the ablation study. On the other hand, we propose a meta concept inference network to infer the task-specific classifier, which is demonstrated to be useful for quickly adapting to a novel task. In future work, we can consider a novel bayesian  inference method for further investigation of the potential on using the concept graph.

%% Loading bibliography style file
%\bibliographystyle{model1-num-names}
\bibliographystyle{cas-model2-names}

% Loading bibliography database
\bibliography{cas-refs}

%\vskip3pt

\end{document}